\documentclass[a4paper, 10pt, conference]{ieeeconf}      % Use this line for a4
\usepackage{FG2021}
\usepackage{xspace}
\usepackage{censor}
\usepackage[nolist]{acronym} % Acronyms from file
\begin{acronym}[PHOENIX14\textbf{T} ]

\acrodefplural{rnn}[RNNs]{Recurrent Neural Networks}
\acrodefplural{cnn}[CNNs]{Convolutional Neural Networks}
\acrodefplural{hmm}[HMMs]{Hidden Markov Models}
\acrodefplural{gru}[GRUs]{Gated Recurrent Units}
\acrodefplural{crf}[CRFs]{Conditional Random Fields}
\acrodefplural{gan}[GANs]{Generative Adversarial Networks}
\acrodefplural{gpu}[GPUs]{Graphic Processing Units}

\acrodefplural{mdn}[MDNs]{Mixture Density Networks}

\acrodefplural{vae}[VAEs]{Variational Autoencoders}

\acro{adain}[AdaIN]{Adaptive Instance Normalisation}

\acro{bsl}[BSL]{British Sign Language}
\acro{bleu}[BLEU]{Bilingual Evaluation Understudy}
\acro{blstm}[BLSTM]{Bidirectional Long Short-Term Memory}
\acro{cnn}[CNN]{Convolutional Neural Network}
\acro{crf}[CRF]{Conditional Random Field}
\acro{cslr}[CSLR]{Continuous Sign Language Recognition}
\acro{ctc}[CTC]{Connectionist Temporal Classification}
\acro{dl}[DL]{Deep Learning}
\acro{dgs}[DGS]{German Sign Language - Deutsche Gebärdensprache}
\acro{dsgs}[DSGS]{Swiss German Sign Language - Deutschschweizer Geb\"ardensprache}
\acro{dtw}[DTW]{Dynamic Time Warping}
\acro{fc}[FC]{Fully Connected}
\acro{ff}[FF]{Feed Forward}
\acro{gan}[GAN]{Generative Adversarial Network}
\acro{gpu}[GPU]{Graphics Processing Unit}
\acro{gru}[GRU]{Gated Recurrent Unit}
\acro{gtpt}[G2PT]{Gloss to Pose Transformer}
\acro{hmm}[HMM]{Hidden Markov Model}
\acro{hpe}[HPE]{Hand Pose Enhancer}
\acro{isl}[ISL]{Irish Sign Language}
\acro{lstm}[LSTM]{Long Short-Term Memory}
\acro{mha}[MHA]{Multi-Headed Attention}
\acro{mtc}[MTC]{Monocular Total Capture}
\acro{mse}[MSE]{Mean Squared Error}
\acro{mdn}[MDN]{Mixture Density Network}
\acro{moe}[MoE]{Mixture-of-Expert}
\acro{nmt}[NMT]{Neural Machine Translation}
\acro{nlp}[NLP]{Natural Language Processing}
\acro{ph12}[PHOENIX12]{RWTH-PHOENIX-Weather-2012}
\acro{ph14}[PHOENIX14]{RWTH-PHOENIX-Weather-2014}
\acro{ph14t}[PHOENIX14\textbf{T}]{RWTH-PHOENIX-Weather-2014\textbf{T}}
\acro{pof}[POF]{Part Orientation Field}
\acro{paf}[PAF]{Part Affinity Field}
\acro{pttt}[P2TT]{Pose to Text Transformer}
\acro{relu}[RELU]{Rectified Linear Units}
\acro{rnn}[RNN]{Recurrent Neural Network}
\acro{rouge}[ROUGE]{Recall-Oriented Understudy for Gisting Evaluation}
\acro{sgd}[SGD]{Stochastic Gradient Descent}
\acro{sla}[SLA]{Sign Language Assessment}
\acro{slr}[SLR]{Sign Language Recognition}
\acro{slt}[SLT]{Sign Language Translation}
\acro{slp}[SLP]{Sign Language Production}
\acro{smt}[SMT]{Statistical Machine Translation}
\acro{slva}[SLVA]{Sign Language Video Anonymisation}

\acro{ttgt}[T2GT]{Text to Gloss Transformer}
\acro{ttpt}[T2PT]{Text to Pose Transformer}
\acro{ttp}[T2P]{Text to Pose}
\acro{ttgtp}[T2G2P]{Text to Gloss to Pose}

\acro{vae}[VAE]{Variational Autoencoder}
\acro{vgt}[VGT]{Flemish Sign Language - Vlaamse Gebarentaal}

\acro{wer}[WER]{Word Error Rate}

\end{acronym} 
\usepackage{graphics} % for pdf, bitmapped graphics files
\usepackage{epsfig} % for postscript graphics files
\usepackage{mathptmx} % assumes new font selection scheme installed
\usepackage{times} % assumes new font selection scheme installed
\usepackage{amsmath} % assumes amsmath package installed
\usepackage{amssymb}  % assumes amsmath package installed

\FGfinalcopy % *** Uncomment this line for the final submission
\IEEEoverridecommandlockouts                              % This command is only
\def\FGPaperID{51} % *** Enter the FG2021 Paper ID here

\ifFGfinal
\newcommand{\datasetname}{Content4All\xspace}
\newcommand{\swisstxt}{SWISSTXT\xspace}
\newcommand{\vrt}{VRT\xspace}
\newcommand{\accesslink}{https://www.cvssp.org/data/c4a-news-corpus}
\newcommand{\toollink}{https://gitlab.com/content4all-public/data-annotation}
\else
\newcommand{\datasetname}{\emph{DATASET\_NAME}\xspace}
\newcommand{\swisstxt}{\emph{BC\_I}\xspace}
\newcommand{\vrt}{\emph{BC\_II}\xspace}
\newcommand{\accesslink}{REMOVED\_FOR\_ANONYMITY\xspace}
\newcommand{\toollink}{REMOVED\_FOR\_ANONYMITY\xspace}
\fi

\title{\LARGE \bf
\datasetname Open Research Sign Language Translation Datasets
}

\author{\parbox{16cm}
    {\centering
        {\large 
            Necati Cihan Camg\"{o}z$^1$ 
            Ben Saunders$^1$ 
            Guillaume Rochette$^1$ \\
            Marco Giovanelli$^2$
            Giacomo Inches$^2$
            Robin Nachtrab-Ribback$^3$ 
            Richard Bowden$^1$
        }\\
        {\normalsize
        $^1$CVSSP, University of Surrey, Guildford, UK
        $^2$Fincons Group, Switzerland
        $^3$SWISSTXT, Switzerland \\
        }
    }
}
\begin{document}

\ifFGfinal
\thispagestyle{empty}
\pagestyle{empty}
\else
\author{Anonymous FG2021 submission\\ Paper ID \FGPaperID \\}
\pagestyle{plain}
\fi
\maketitle

%%%%%%%%%%%%%%%%%%%%%%%%%%%%%%%%%%%%%%%%%%%%%%%%%%%%%%%%%%%%%%%%%%%%%%%%%%%%%%%%
\begin{abstract}
Computational sign language research lacks the large-scale datasets that enables the creation of useful real-life applications. To date, most research has been limited to prototype systems on small domains of discourse, e.g. weather forecasts. To address this issue and to push the field forward, we release six datasets comprised of 190 hours of footage on the larger domain of news. From this, 20 hours of footage have been annotated by Deaf experts and interpreters and is made publicly available for research purposes. In this paper, we share the dataset collection process and tools developed to enable the alignment of sign language video and subtitles, as well as baseline translation results to underpin future research.
\end{abstract}

%%%%%%%%%%%%%%%%%%%%%%%%%%%%%%%%%%%%%%%%%%%%%%%%%%%%%%%%%%%%%%%%%%%%%%%%%%%%%%%%
\section{Introduction}

Sign languages are natural visual languages \cite{sutton1999linguistics}. They are the main medium of communication of the Deaf and have rich grammatical rules with unique syntax and vocabulary. Signers utilize multiple complementary articulators to convey information via spatio-temporal constructs \cite{boyes2001hands}. Machine understanding of sign languages is a challenging task and an active research field \cite{bragg2019sign}, that has seen recent progress with the availability of benchmark recognition \cite{forster2014extensions,pu2019iterative,potamianos2020sl} and translation \cite{camgoz2018neural,adaloglou2020comprehensive} datasets. 

As with any machine translation task, sign language translation \cite{camgoz2020multi} and production \cite{saunders2020sign} require large-scale corpora to build models that generalize effectively. However, large-scale datasets with high quality annotations, on par with their spoken language counterparts, are not available for computational sign language research. This not only limits the advancement of the field, but also gives the research and Deaf communities a false sense of technological readiness by reporting promising results on small datasets with limited domains of discourse.

To address these issues, in this work we introduce 190 hours of public datasets, an order of magnitude larger than currently available resources, aimed at supporting sign language translation research \cite{camgoz2018neural}. In collaboration with broadcast partners, we collected sign language interpretation footage on the large domain of news. We adapted an annotation tool \cite{dutta2019vgg} which was used to manually align spoken language subtitles to sign language videos\footnote{\toollink}. We extracted human body pose features from all the videos and share them as part of the dataset to reduce redundant compute. Additionally, we trained state-of-the-art transformer based translation models \cite{camgoz2020multi} using the pose features and report baseline results. The datasets are publicly available\footnote{\accesslink} for research purposes upon agreeing to a user license (CC 4.0 BY-NC-SA\footnote{https://creativecommons.org/licenses/by-nc-sa/4.0/}).\looseness=-1

\begin{figure}[t!]
    \centering
    \includegraphics[width=0.92\linewidth]{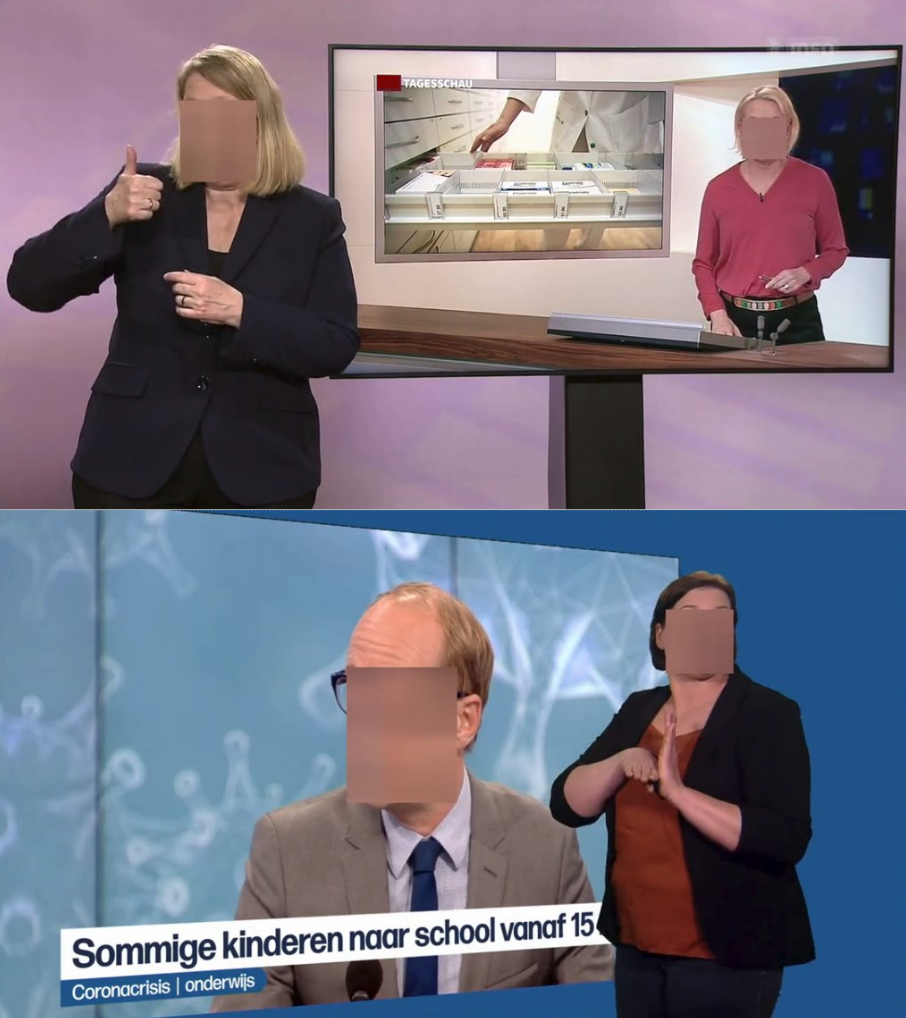}
    \caption{Anonymized video frame samples from \swisstxt (top) and \vrt (bottom) broadcast footage}
    \label{fig:anon:sample}
\end{figure}

The rest of this paper is structured as follows: In Section~\ref{sec:datasets} we give a brief (non-exhaustive) overview of existing datasets that have been used for computational sign language research. We introduce the \datasetname base datasets in Section~\ref{sec:raw:dataset}. We then give details of the subset selection and annotation process, and introduce the manually aligned datasets in Section~\ref{sec:annotated:dataset}. We report baseline translation performance in Section~\ref{sec:experiments} to underpin future research. Finally, we conclude this paper and discuss possible future directions in Section~\ref{sec:conclusions}.

% %%%%%%%%%%%%%%%%%%%%%%%%%%%%%%%%%%%%%%%%%%%%%%%%%%%%%%%%%%%%%%%%%%%%%%%%%%%%%%%
\begin{table*}[]
\caption{Detailed statistics of the released three base (RAW) datasets.}
\label{tbl:raw}
\begin{center}
\begin{tabular}{l|l|l|l}
        & \swisstxt-RAW-NEWS & \swisstxt-RAW-WEATHER & \vrt-RAW  \\ \hline
Sign Interpretations & \acs{dsgs} & \acs{dsgs} &\acs{vgt} \\
Subtitle Language    & Swiss-German & Swiss-German & Flemish \\
Number of Videos     & 183  & 181 & 120 \\
Duration (hours:minutes) & 76:05 & 12:09 & 100:21 \\
Video Resolution@FPS & 1280x720@50fps & 1280x720@50fps & 1280x720@25fps \\
Video Codec/Container  & H264/mp4 & H264/mp4 & H264/mp4 
\end{tabular}
\end{center}
\end{table*}
% %%%%%%%%%%%%%%%%%%%%%%%%%%%%%%%%%%%%%%%%%%%%%%%%%%%%%%%%%%%%%%%%%%%%%%%%%%%%%%%

\section{Related Work}
\label{sec:datasets}

Compared to their spoken counterparts, computational sign language research is an under-resourced field, especially with regards to data availability \cite{bragg2019sign}. Until recently, most computational sign language research was focused on isolated sign recognition \cite{camgoz2016sign,cooper2007sign,joze2019ms}, working on datasets collected in controlled environments with a limited vocabulary \cite{ebling2018smile,camgoz2016bosphorussign,chai2014devisign}. This resulted in most research being incomparable whilst also robbing the field of competitive progress \cite{koller2020quantitative}.

The available large-scale linguistic resources, such as BSLCP \cite{schembri2013building} and mein-DGS \cite{hanke2010dgs}, have been considered as alternate resources for sign language recognition \cite{cormier2019extol}. However, they were not collected with vision-based sign language research in mind. They lack the necessary inter- and intra- signer variance on shared content, which current approaches rely on to build generalizable statistical models. 

Relatively recently, the field has shifted focus to continuous recognition of co-articulated sign sequences \cite{koller2015continuous}. To support this move, researchers have collected datasets based on sentence repetition, where signers were asked to repeat preconstructed sign phrases \cite{pu2019iterative, adaloglou2020comprehensive, potamianos2020sl}. Although these datasets are useful for their target application, such as sign language assessment \cite{tornay2019hmm,aran2009signtutor,zafrulla2011american}, they have a limited vocabulary and phrase variance, thus making them less useful for building large-scale sign language translation models.

Sign interpretations from broadcast footage have also been considered as alternate sources of data \cite{cooper2009learning,pfister2012automatic}. Given the variety and quantity of the signed content, they are an ideal resource for training sign language translation models that generalize to large domains of discourse. A good example of broadcast footage utilization for computational sign language research is the Phoenix corpora \cite{forster2012rwth, forster2014extensions}. Comprised of sign language interpretations from broadcast weather forecast footage, the Phoenix datasets have been widely used for continuous sign language recognition research \cite{zhou2020spatialtemporal, koller2019weakly, camgoz2017subunets}. With the release of the aligned spoken language translations \cite{camgoz2018neural}, Phoenix2014T has become the benchmark dataset for sign language translation. Although researchers have developed successful models which achieve promising results on Phoenix2014T \cite{camgoz2020multi, camgoz2020sign, yin2020better}, the application of such models has been limited to the weather domain, due to the limited domain of discourse of the dataset. Furthermore, compared to contemporary broadcast technology (1280x720@50fps), the dataset is of poor visual quality. It has low image resolution (210x260) and suffer from motion blur and interlacing artifacts, due to being recorded between the period of 2009-2013.\looseness=-1

To address the shortcomings of the available public datasets, in this work we release six new sign language translation datasets of broadcast sign language interpretations. We share an order of magnitude larger amount of data compared to the benchmark Phoenix2014T dataset \cite{camgoz2018neural} (190 hours vs 10 hours, respectively) in two different sign languages, namely \ac{dsgs} and \ac{vgt}. We provide both sign language interpretations in the large domain of news broadcasts and the corresponding spoken language subtitles for each program, as well as manual spoken language text-sign video alignments for three of the datasets.

\section{\datasetname Base (RAW) Datasets}
\label{sec:raw:dataset}

In this section we introduce the three base datasets that have been created using the broadcast sign language interpretation footage. Broadcast partners \swisstxt and \vrt recorded and provided news footage from the periods March-August 2020 and March-July 2020, respectively. All of the video footage contains picture-in-picture sign language interpretations in \ac{dsgs} for \swisstxt and \ac{vgt} for \vrt.

In total, \swisstxt and \vrt provided 183 and 120 days of news footage, yielding 152 hours 40 minutes and 100 hours 21 minutes of video data, respectively. Both sources consist of mp4 files with a resolution of 1280x720. Both broadcasters provided time-coded subtitles in their respective spoken language, which we also make publicly available.

% %%%%%%%%%%%%%%%%%%%%%%%%%%%%%%%%%%%%%%%%%%%%%%%%%%%%%%%%%%%%%%%%%%%%%%%%%%%%%%%
\begin{table*}[]
\caption{Detailed statistics of the three manually aligned datasets in contrast to Phoenix2014T.}
\label{tbl:aligned}
\begin{center}
\begin{tabular}{l|l|l|l|l}
    & Phoenix2014T & \swisstxt-WEATHER & \swisstxt-NEWS & \vrt-NEWS  \\ \hline
\#sequences          & 8257 & 811 & 6031 & 7174 \\
\#frames             & 947,756 & 156,891 & 1,693,391 & 810,738      \\
Duration (hour:mins:secs) & 10:31:50 & 00:52:18 & 09:24:28 & 09:00:30     \\
Total Words (TW)     & 99,081 & 6,675 & 72,892 & 79,833 \\
Vocabulary Size (VS) & 2,287  & 1,248 & 10,561 & 6,875 \\
NS (/VS)             & 1,077 (0.37) & 701 (0.56) & 5,969 (0.57) & 3,405 (0.50) \\
\textless{} 3 (/VS)  & 1,427 (0.49) & 874 (0.70) & 7,561 (0.72) & 4423 (0.64)  \\
\textless{} 5 (/VS)  & 1,758 (0.60) & 999 (0.80) & 8,779 (0.83) & 5334 (0.78)  \\
\textless{}10 (/VS)  & 2,083 (0.72) & 1,124 (0.90) & 9,661 (0.91) & 6009 (0.87) 
\end{tabular}
\vspace{0.1in}
\end{center}
\end{table*}
% %%%%%%%%%%%%%%%%%%%%%%%%%%%%%%%%%%%%%%%%%%%%%%%%%%%%%%%%%%%%%%%%%%%%%%%%%%%%%%%

\vrt videos contain uninterrupted sign language interpretations (\vrt-RAW). However, \swisstxt broadcast footage is interlaced with advertisements, which are not signed. We have manually annotated and removed the advertisement regions, creating two new datasets, namely \swisstxt-RAW-NEWS and \swisstxt-RAW-WEATHER. The detailed statistics of all the base (RAW) datasets can be seen in Table~\ref{tbl:raw}.

To encourage the use of the datasets and to reduce redundant compute, we share extracted 2D and 3D human pose estimation features for all the video frames. We first ran the BODY-135 variant of the OpenPose library \cite{hidalgo2019single} and obtained 2D human pose information. The obtained 2D body skeletons are then uplifted using a deep learning based 2D-to-3D uplifting method proposed in \cite{rochette2019weakly}. Both 2D and 3D pose information are stored in json files, with one file per frame. We compress all json files corresponding to the same video into a single tar.xz file, for the sake of storage ease and efficiency. We also share a sample Python script to access them efficiently.

All of the footage has been anonymized to protect the privacy of the individuals present in the released datasets. We utilize the facial landmark predictions of OpenPose to locate each individual’s face and blur the corresponding regions using a Gaussian kernel. Anonymized video frame samples can be seen in Figure~\ref{fig:anon:sample}. Interested parties are advised to contact the broadcast partners for the non-anonymized versions of the video files.

In summary, the curated base (RAW) datasets include:
\begin{itemize}
    \item Anonymized news footage with picture-in-picture sign language interpretations.
    \item Corresponding spoken language subtitles.
    \item Human pose estimation features, namely 2D OpenPose \cite{hidalgo2019single} and 3D skeleton \cite{rochette2019weakly} outputs.
\end{itemize}

\section{\datasetname Manually Aligned Datasets}
\label{sec:annotated:dataset}

Automatic sign language translation has so far only been successfully realized for a small domain of discourse, namely weather forecasts using the Phoenix2014T dataset~\cite{camgoz2018neural}. However, the same techniques fail to learn successful translation models on larger domains of discourse using similar amounts of data \cite{cormier2019extol}.

To move the field beyond the weather domain, whilst also considering the aforementioned limitations of learning-based translation techniques, we selected 10 hour subsets from both the \swisstxt-RAW-NEWS and \vrt-RAW datasets that best resembles the Phoenix2014T dataset statistics. We manually aligned spoken language text and sign language video pairs in these subsets with the help of Deaf experts and sign language interpreters. These two new datasets, \swisstxt-NEWS and \vrt-NEWS, contain over 13,000 sign language interpretation and spoken language translation pairs. Whilst building up the annotation process, we also annotated a subset of the \swisstxt-RAW-WEATHER dataset, which we also make publicly available for research purposes (\swisstxt-WEATHER). 

The detailed statistics of all three manually annotated datasets in comparison to Phoenix2014T can be seen in Table~\ref{tbl:aligned}. We calculate the total number of words, vocabulary size and number of singletons (words that only occur once). We also share the number of unique words that occur less than three (\textless{}3), five (\textless{}5) and ten (\textless{}10) times. To enable a comparison between the different datasets, we also report metrics proportional to the vocabulary size (/VS).

Similar to the base (RAW) datasets (Section~\ref{sec:raw:dataset}), we also provide anonymized sign language videos and 2D \& 3D human body pose estimates for these annotated datasets. In addition, we provide a json file for each dataset, containing the spoken language sentence-sign language interpretation alignments. Each element in the json files has a unique ID that corresponds to a spoken language sentence-sign phrase pair. Alongside the spoken language translations, the beginning and end timing information in \textit{minutes:seconds:milliseconds} and in frames are provided, which indicates the location of the sign video-sentence alignment in the corresponding video. Within each element, we also store the relative path of corresponding videos and the 2D/3D skeletal files. 

In the rest of this section we give further details on the subset selection and manual alignment processes.

\subsection{Subset Selection}

We first parsed the subtitles and manually cleaned the text to eliminate special characters, i.e. replaced periods (``.'') which are not used to terminate sentences with a special token. We then tokenized subtitle sentences into words, and calculated word statistics. To determine the size of the domain of discourse, we used spoken language word statistics as a proxy. 

To select 10 hour subsets for manual subtitle alignment, we implemented a beam search algorithm. In order to constrain the domain of discourse, we capitalised on the COVID focused news during the capture period and used the first 22 minutes of each video. Considering some parts of the videos are eliminated during the annotation process, we selected 30 videos for each dataset, thus yielding a total of 11 hours of footage for each dataset. We used a beam size of 50 and use vocabulary size as the key metric to select a combination of videos that have similar statistics to the Phoenix2014T corpus. The selected subset was then annotated using the annotation tool described in the following section.

\subsection{Manual Alignment}

Current sign language translation techniques require parallel corpora, which provide spoken language translations that correspond to segmented sign language videos. The curation of such a dataset is a challenging task, as spoken and sign languages do not share a one-to-one word mapping and have a non-monotonic relationship. We utilized subtitles obtained from broadcast footage and aligned these subtitles to sign language interpretations. 

Even though there are sign language video annotation tools already available (e.g. ELAN \cite{wittenburg2006elan}), they mainly focus on data entry for linguistic purposes, such as the annotation of glosses. In our specific use case, we need to manually adjust weakly pre-aligned sentences (based on subtitles) with the corresponding sign language interpreter video in order to speed up the annotation work.

\begin{figure}[t!]
    \centering
\ifFGfinal
    \includegraphics[width=0.82\linewidth]{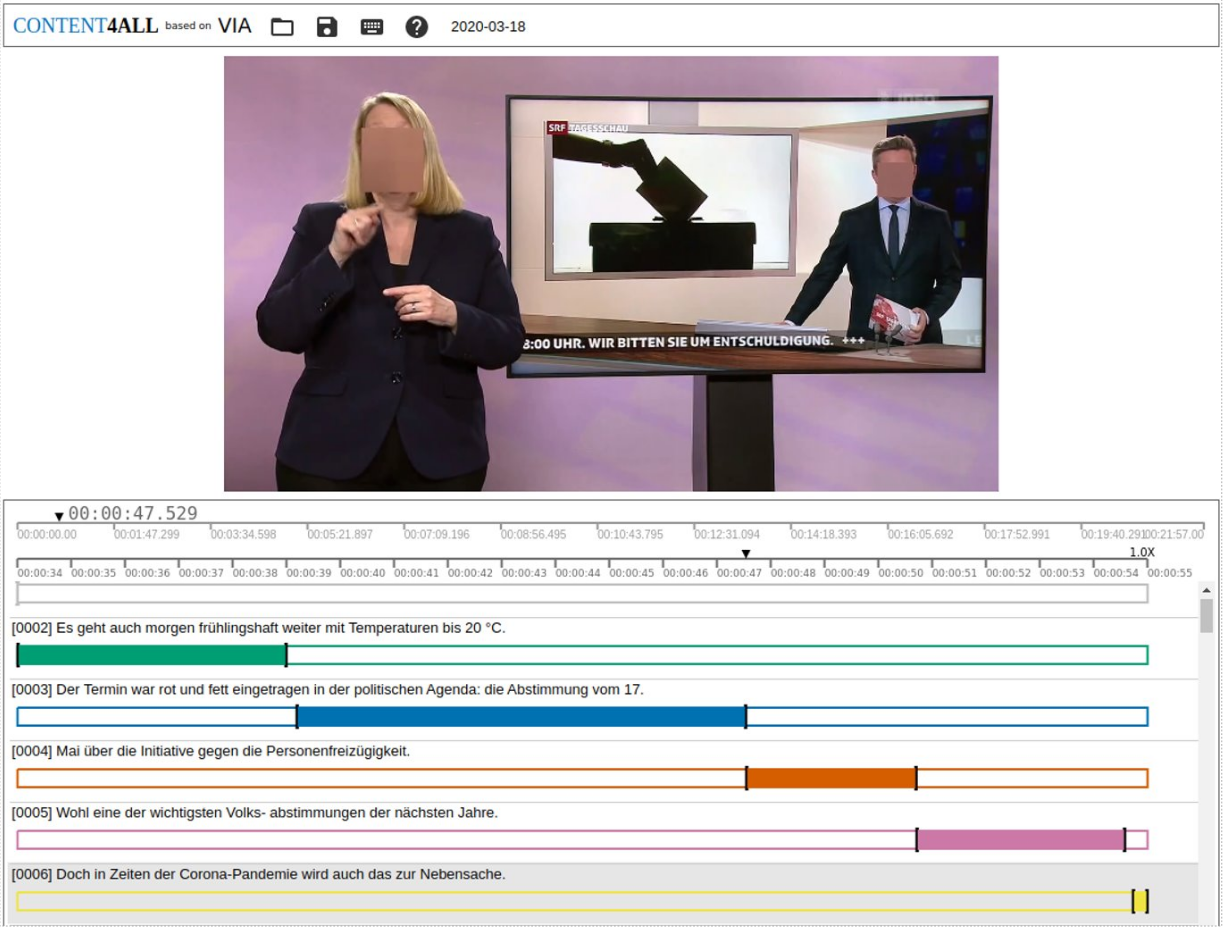}
\else
    \includegraphics[width=0.82\linewidth]{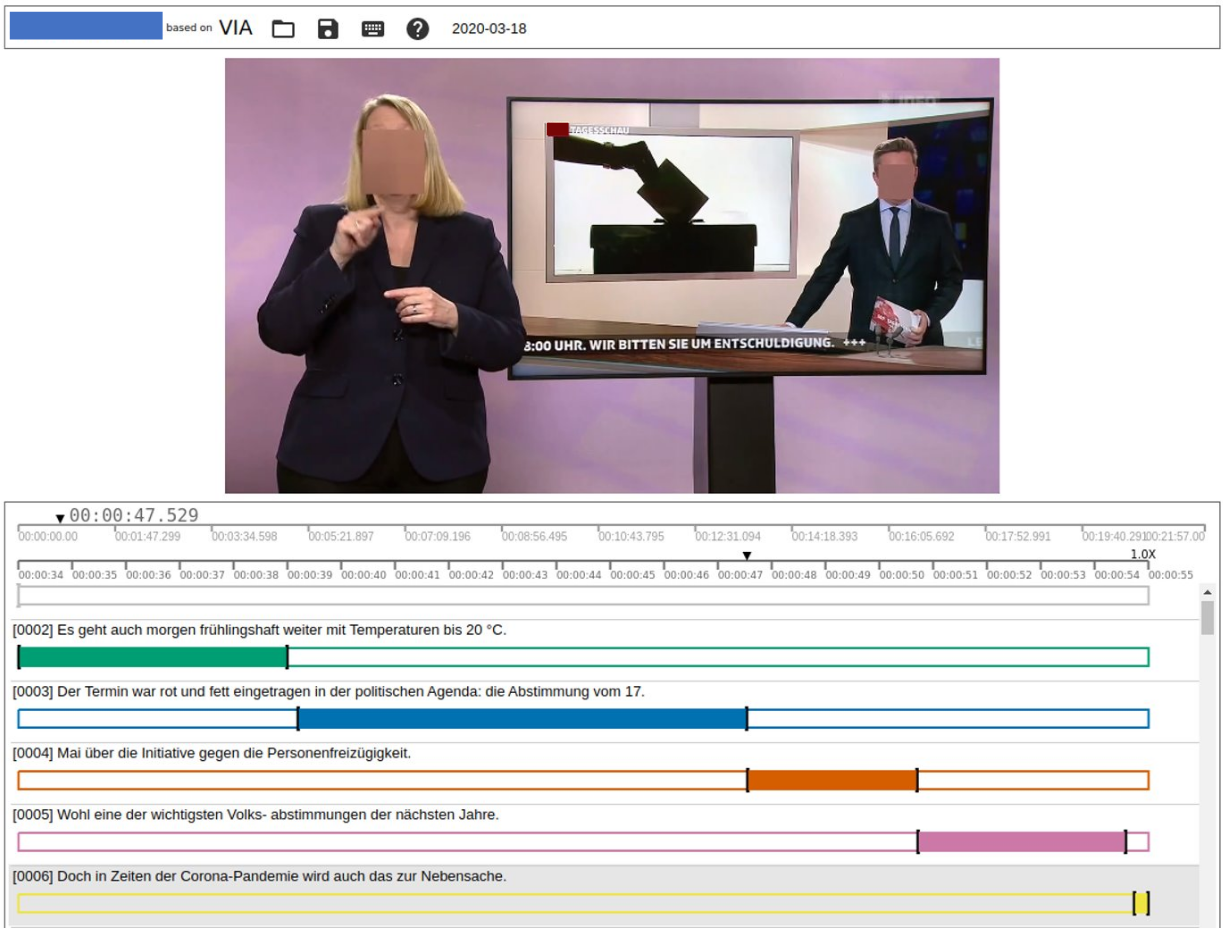}
\fi
    \caption{Sign language interpretation - subtitle alignment tool user interface.}
    \label{fig:annotation:tool}
\end{figure}

We used the VIA annotation tool \cite{dutta2019vgg} developed by the University of Oxford, and adapted it to facilitate the alignment of spoken language subtitles to sign language videos. We converted the subtitle files into VIA project file format. The project files were then sent to the Deaf experts and interpreters via email with a link to the annotation platform. The annotators were asked to align spoken language text to sign language videos when applicable. An equivalent of three person-months was needed to annotate 11 hours of video in each sign language. The user interface of the modified tool used by the annotators can be seen in Figure~\ref{fig:annotation:tool}. 

\section{Baseline Experiments}
\label{sec:experiments}

To underpin future research on the released datasets, we train sign language translation models and report baseline results on \swisstxt-NEWS and \vrt-NEWS.

\textbf{Protocols:} To make future research on our datasets comparable, we set training and testing protocols. We divide each dataset into three partitions, namely training, development and test set. Detailed partition statistics for \swisstxt-NEWS and \vrt-NEWS can be seen in Table \ref{tbl:protocol}.

\begin{table}[!hb]
\caption{statistics of the evaluation protocols.} \label{tbl:protocol}
\begin{center}
\resizebox{0.48\textwidth}{!}{
\begin{tabular}{l|lll|lll}
           & \multicolumn{3}{c|}{\swisstxt-NEWS} & \multicolumn{3}{c}{\vrt-NEWS} \\
                 & Train     & Dev     & Test    & Train     & Dev    & Test \\ \hline
\#segments       & 5,126     & 442     & 463     & 6,098     & 542    & 534  \\
\#frames         & 1,445,804 & 122,666 & 130,952 & 698,077   & 60,887 & 58,948 \\
Total Word Count & 62,302    & 5,382   & 5,774   & 67,973    & 6,060  & 5,858  \\
Vocabulary Size  & 9,623     & 1,976   & 2,093   & 6,325     & 1,612  & 1,584  \\
\#Out-of-Vocab   & --        & 455     & 500     &           & 304    & 257    \\
\#Singletons     & 5,535     & --      & --      & 3,190     & --     & --       
\end{tabular}
}
\end{center}
\end{table}

To measure the translation performance, we use \ac{rouge} \cite{lin2004rouge} and \ac{bleu} \cite{papineni2002bleu} scores, which are the most common metrics used for machine and sign language translation \cite{camgoz2020multi,camgoz2020sign,saunders2020adverserial}.

\textbf{Experimental Setup and Results:} We employ a state-of-the-art transformer~\cite{vaswani2017attention} based sign language translation approach~\cite{camgoz2020sign}. We represent video frames with the extracted 2D human pose features and use them as inputs to our models. We construct our transformers using three layers, with 512 hidden units, 1024 feed forward size and 8 heads per layer. We use Xavier initilization \cite{glorot2010understanding} and train all our networks from scratch. We also utilize dropout with a drop rate of 0.3 to help our models generalize.\looseness=-1

We use Adam \cite{kingma2014adam} optimizer with a batch size of 32 and a learning rate of $10^{-3}$. We employ a plateau learning rate scheduling algorithm which tracks the development set performance. We evaluate our network every 100 iterations. If the development set performance does not improve for 8 evaluation cycles, we decrease the learning rate by a factor of 0.7. This continues until the learning rate drops below $10^{-5}$.\looseness=-1

During inference, we utilize a beam search decoding with a length penalty \cite{wu2016google}. We determine the best performing beam width and length penalty $\alpha$ values based on development set performance. We then use these values for reporting results on the test set.

\begin{table}[!h]
\caption{Baseline results on our datasets.} \label{tbl:results}
\begin{center}
\begin{tabular}{l|cc|cc}
              & \multicolumn{2}{c}{Dev} & \multicolumn{2}{|c}{Test} \\
              & ROGUE      & BLEU-4     & ROGUE      & BLEU-4      \\ \hline
\swisstxt-NEWS & 15.72     & 0.46       & 15.34    & 0.41     \\
\vrt-NEWS      & 17.63     & 0.45       & 17.77    & 0.36            
\end{tabular}
\end{center}
\end{table}

We achieve similar baseline translation performances on both datasets (See Table~\ref{tbl:results}). These results not only display the challenge of moving to larger domains of discourse, but also encourage the development of more specialized sign language translation approaches to realize real-life computational sign language applications. 

\section{Conclusions}
\label{sec:conclusions}

In this paper we introduced six new open research datasets aimed at automatic sign language translation research. Sign language interpretation footage was captured by the broadcast partners \swisstxt and \vrt. Raw footage was anonymized and processed to extract 2D and 3D human body pose information. From the roughly 190 hours of processed data, three base (RAW) datasets were released, namely 1)~\swisstxt-RAW-NEWS, 2)~\swisstxt-RAW-WEATHER and 3)~\vrt-RAW. Each dataset contains sign language interpretations, corresponding spoken language subtitles, and extracted 2D/3D human pose information.

A subset from each base dataset was selected and manually annotated to align spoken language subtitles and sign language interpretations. The subset selection was done to resemble the benchmark Phoenix 2014T dataset. Our aim is for these three new annotated public datasets, namely 4)~\swisstxt-NEWS, 5)~\swisstxt-WEATHER and 6)~\vrt-NEWS to become benchmarks and underpin future research as the field moves closer to translation and production on larger domains of discourse.

%%%%%%%%%%%%%%%%%%%%%%%%%%%%%%%%%%%%%%%%%%%%%%%%%%%%%%%%%%%%%%%%%%%%%%%%%%%%%%%%
\ifFGfinal
\small{
\section*{Acknowledgements}
We would like to thank Gregg Young (VRT), Harry Witzthum and Swiss Federation of the Deaf for their contributions to the dataset creation process. This work received funding from the European Union's Horizon2020 research and innovation programme under grant agreement no. 762021 `Content4All'. This work reflects only the authors view and the Commission is not responsible for any use that may be made of the information it contains.
}
\fi
%%%%%%%%%%%%%%%%%%%%%%%%%%%%%%%%%%%%%%%%%%%%%%%%%%%%%%%%%%%%%%%%%%%%%%%%%%%%%%%%

\newpage

{\small
\bibliographystyle{ieee}
\bibliography{_Bib/action,_Bib/camgoz,_Bib/ctc,_Bib/deeplearning,_Bib/generative,_Bib/gesture,_Bib/misc,_Bib/nmt,_Bib/pose,_Bib/seq2seq,_Bib/sign,_Bib/speech}
}

\end{document}